# Induction of Selective Bayesian Classifiers

PAT LANGLEY (LANGLEY@FLAMINGO.STANFORD.EDU)
STEPHANIE SAGE (SAGE@FLAMINGO.STANFORD.EDU)
Institute for the Study of Learning and Expertise
2451 High Street, Palo Alto, CA 94301

## Abstract

In this paper, we examine previous work on the naive Bayesian classifier and review its limitations, which include a sensitivity to correlated features. We respond to this problem by embedding the naive Bayesian induction scheme within an algorithm that carries out a greedy search through the space of features. We hypothesize that this approach will improve asymptotic accuracy in domains that involve correlated features without reducing the rate of learning in ones that do not. We report experimental results on six natural domains, including comparisons with decision-tree induction, that support these hypotheses. In closing, we discuss other approaches to extending naive Bayesian classifiers and outline some directions for future research.

## Introduction

In recent years, there has been growing interest in probabilistic methods for induction. Such techniques have a number of clear attractions: they accommodate the flexible nature of many natural concepts; they have inherent resilience to noise; and they have a solid grounding in the theory of probability. Moreover, experimental studies of probabilistic methods have revealed behaviors that are often competitive with the best inductive learning schemes.

Although much of the recent work on probabilistic induction (e.g., Anderson & Matessa, 1992; Cheeseman et al., 1988; Fisher, 1987; Hadzikadic & Yun, 1989; McKusick & Langley, 1991) has focused on unsupervised learning, the same basic approach applies equally well to supervised learning tasks. Supervised Bayesian methods have long been used within the field of pattern recognition (Duda & Hart, 1973), but only in the past few years have they received attention within the machine learning community (e.g., Clark & Niblett, 1989; Kononenko, 1990, 1991; Langley, Iba, & Thompson, 1992).

In this paper we describe a technique designed to improve upon the already impressive behavior of the simplest approach to probabilistic induction – the naive Bayesian classifier. Below we review the representational, performance, and learning assumptions that underlie this method, along with some situations in which they can lead to problems. One central assumption made by the naive approach is that attributes are independent within each class, which can harm the classification process when violated.

In response to this drawback, we describe a revised algorithm – the selective Bayesian classifier – that deals with highly correlated features by incorporating only some attributes into the final decision process. We present experimental evidence that this scheme improves asymptotic accuracy in domains where the naive classifier fares poorly, without hurting behavior in domains where it compares to other induction algorithms. We close the paper with some comments on related work and directions for future research.

## The Naive Bayesian Classifier

The most straightforward and widely tested method for probabilistic induction is known as the *naive Bayesian classifier*.[1] This scheme represents each class with a single probabilistic summary. In particular, each description has an associated class probability or base rate, $p(C_k)$, which specifies the prior probability that one will observe a member of class $C_k$. Each description also has an associated set of conditional probabilities, specifying a probability distribution for each attribute. In nominal domains, one typically stores a discrete distribution for each attribute in a description. Each $p(v_j|C_k)$ term specifies the probability of value $v_j$, given an instance of class $C_k$. In numeric domains, one must represent a continuous probability distribution for each attribute. This requires that one assume some general form or model, with a common choice being the normal distribution, which can be conveniently represented entirely in terms of its mean and variance.

---

1. We have borrowed this term from Kononenko (1990); other common names for the method include the *simple Bayesian classifier* (Langley, 1993) and *idiot Bayes* (Buntine, 1990).

To classify a new instance $I$, a naive Bayesian classifier applies Bayes' theorem to determine the probability of each description given the instance,

$$p(C_i|I) = \frac{p(C_i)p(I|C_i)}{p(I)} \quad .$$

However, since $I$ is a conjunction of $j$ values, one can expand this expression to

$$p(C_i|\bigwedge v_j) = \frac{p(C_i)p(\bigwedge v_j|C_i)}{\sum_k p(\bigwedge v_j|C_k)p(C_k)} \quad ,$$

where the denominator sums over all classes and where $p(\bigwedge v_j|C_i)$ is the probability of the instance $I$ given the class $C_i$. After calculating these quantities for each description, the algorithm assigns the instance to the class with the highest probability.

In order to make the above expression operational, one must still specify how to compute the term $p(\bigwedge v_j|C_k)$. The naive Bayesian classifier assumes independence of attributes within each class, which lets it use the equality

$$p(\bigwedge v_j|C_k) = \prod_j p(v_j|C_k) \quad ,$$

where the values $p(v_j|C_k)$ represent the conditional probabilities stored with each class. This approach greatly simplifies the computation of class probabilities for a given observation.

The Bayesian framework also lets one specify prior probabilities for both the class and the conditional terms. In the absence of domain-specific knowledge, a common scheme makes use of 'uninformed priors', which assign equal probabilities to each class and to the values of each attribute. However, one must also specify how much weight to give these priors relative to the training data. For example, Anderson and Matessa (1992) use a Dirichlet distribution to initialize probabilities and give these priors the same influence as a single training instance. Clark and Niblett (1989) describe another approach that does not use explicit priors, but instead estimates $P(C_k)$ and $p(v_j|C_k)$ directly from their proportions in the training data. When no instances of a value have been observed, they replace the zero probability with $p(C_i)/N$, where $N$ is the number of training cases.[2]

Learning in the naive Bayesian classifier is an almost trivial matter. The simplest implementation increments a count each time it encounters a new instance, along with a separate count for a class each time it observes an instance of that class. These counts let the classifier estimate $p(C_k)$ for each class $C_k$. For each nominal value, the algorithm updates a count for that class-value pair. Together with the second count, this lets the classifier estimate $p(v_j|C_k)$. For each numeric attribute, the method retains and revises two quantities, the sum and the sum of squares, which let it compute the mean and variance for a normal curve that it uses to find $p(v_j|C_k)$. In domains that can have missing attributes, it must include a fourth count for each class-attribute pair.

In contrast to many induction methods, the naive Bayesian classifier does not carry out an extensive search through a space of possible descriptions. The basic algorithm makes no choices about how to partition the data, which direction to move in a weight space, or the like, and the resulting probabilistic summary is completely determined by the training data and the prior probabilities. Nor does the order of the training instances have any effect on the output; the basic process produces the same description whether it operates incrementally or nonincrementally. These features make the the learning algorithm both simple to understand and quite efficient.

Bayesian classifiers would appear to have advantages over many induction algorithms. For example, their collection of class and conditional probabilities should make them inherently robust with respect to noise. Their statistical basis should also let them scale well to domains that involve many irrelevant attributes. Langley, Iba, and Thompson (1992) present an average-case analysis of these factors' effect on the algorithm's behavior for a specific class of target concepts.

The experimental literature is consistent with these expectations, with researchers reporting that the naive Bayesian classifier gives remarkably high accuracies in many natural domains. For example, Cestnik, Kononenko, and Bratko (1987) included this method as a straw man in their experiments on decision-tree induction, but found that it fared as well as the more sophisticated techniques. Clark and Niblett (1989) reported similar results, finding that the naive Bayesian classifier learned as well as both rule-induction and decision-tree methods on medical domains. And Langley et al. (1992) obtained even stronger results, in which the simple probabilistic method outperformed a decision-tree algorithm on four out of five natural domains.

However, the naive Bayesian classifier relies on two important assumptions. First, this simple scheme posits that the instances in each class can be summarized by a single probabilistic description, and that these are sufficient to distinguish the classes from one other. If we represent each attribute value as a feature that may be present or absent, this is closely related to the assumption of linear separability in early work on neural networks. Other encodings lead to a more complex story, but the effect is nearly the same. Nevertheless, like perceptrons, Bayesian classifiers are typically limited to learning classes that can be separated by a single decision boundary.[3] Although we have addressed this limitation in other work (Langley, 1993), we will not focus on it here.

---

2. This technique has no solid basis in probability theory, but it avoids arbitrary parameters and it approximates other approaches after only a few instances; thus, we have used it in our implementations.

Another important assumption that the naive Bayesian classifier makes is that, within each class, the probability distributions for attributes are independent of each other. One can model attribute dependence within the Bayesian framework (Pearl, 1988), but determining such dependencies and estimating them from limited training data is much more difficult. Thus, the independence assumption has clear attractions. Unfortunately, it is unrealistic to expect this assumption to hold in the natural world. Correlations among attributes in a given domain are common. For example, in the domain of medical diagnosis, certain symptoms are more common among older patients than younger ones, regardless of whether they are ill. Such correlations introduce dependencies into the probabilistic summaries that can degrade a naive Bayesian classifier's accuracy.

To illustrate this difficulty, consider the extreme case of redundant attributes. For a domain with three features, the numerator we saw earlier becomes

$$p(C_i)p(v_1|C_i)p(v_2|C_i)p(v_3|C_i) \quad .$$

If we include a fourth feature that is perfectly correlated (redundant) with the first of these features, we obtain

$$p(C_i)p(v_1|C_i)^2 p(v_2|C_i)p(v_3|C_i) \quad ,$$

in which $v_1$ has twice as much influence as the other values. The emphasis given to the redundant information reduces the influence of other features, which can produce a biased prediction. For example, consider a linearly separable target concept that predicts class $A$ is any two of three features are present and that predicts class $B$ otherwise. A naive classifier can easily master this concept, but given a single redundant feature, it will consistently misclassify one of the eight possible instances no matter how many training cases it encounters.

Surprisingly, many of the domains in which the naive Bayesian classifier performs well appear to contain significant dependencies. This evidence comes in part from Holte's (1993) studies, which show that one-level decision trees do nearly as well as full decision trees on many of these domains. In addition, Langley and Sage (1994) found that the behavior of a simple nearest neighbor algorithm does not suffer in these domains, as one would expect if there were many irrelevant attributes. Since one attribute is sufficient for high accuracy and the remaining ones do not degrade a nearest neighbor method, then many of the attributes would appear to be highly correlated.

The strong performance of the naive Bayesian method despite violation of the independence assumption is intriguing. It suggests that a revised method which circumvents dependencies should outperform the naive algorithm in domains where dependencies occur, while performing equally well in cases where they do not. In the following section, we discuss a variant Bayesian algorithm that selects and uses a subset of the known features in an attempt to exclude highly correlated attributes. This should let one continue to make the convenient assumption of independence while minimizing its detrimental effects on classification accuracy.

## The Selective Bayesian Classifier

Our goal was to modify the naive Bayesian classifier to achieve improved accuracy in domains with redundant attributes. The selective Bayesian classifier is a variant of the naive method that uses only a subset of the given attributes in making predictions. In other words, the performance component of the algorithm computes $p(\bigwedge v_j|C_k)$ as the product of conditional probabilities, $p(v_j|C_k)$, for selected attributes $v_j$ from the original feature set. The learning component of the selective classifier augments the original algorithm with the ability to exclude attributes that introduce dependencies. This process consists of a search through the space of attribute subsets.

We made a number of choices in designing the search process. First, the direction of search could proceed in a forward or backward manner. A forward selection method would start with the empty set and successively add attributes, while a backward elimination process would begin with the full set and remove unwanted ones. A potential problem with backward search is that, when several attributes are correlated, removing any one of them may not improve performance since redundant information will still exist. We chose to use forward selection since it should immediately detect dependencies when a harmful redundant attribute is added.

A second decision dealt with the organization of the search. Clearly, an exhaustive search of the space is impractical, since there are $2^a$ possible subsets of $a$ attributes. A more realistic approach, commonly used in machine learning algorithms, is to use a greedy method to traverse the space. That is, at each point in the search, the algorithm considers all local changes to the current set of attributes, makes its best selection, and never reconsiders this choice. This gives a worst-case time complexity of $O(a^2)$.

---

3. The main exception involves numeric domains; Duda and Hart (1973) present a simple situation in which two decision boundaries emerge from the use of normal distributions.

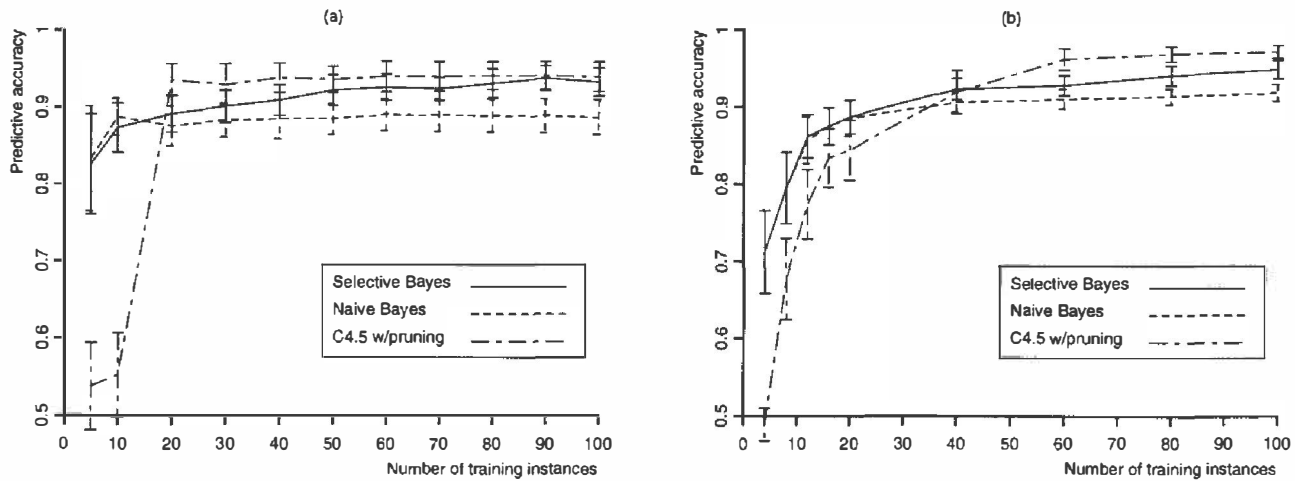

*Figure 1*. Learning curves for the selective Bayesian classifier, the naive Bayesian classifier, and C4.5 with pruning on (a) Congressional voting records and (b) the mushroom domain. The error bars represent 95% confidence intervals based on a two-sided $t$ test.

Third, we needed some metric to evaluate alternative subsets of attributes. We considered the leave-one-out technique for estimating accuracy from the training set, since this is the most accurate method of cross validation. Moreover, it can be applied efficiently to a Bayesian classifier since one can simply 'subtract' a given instance from the stored attribute frequencies, measure the accuracy of the resulting classifier, and add the instance back. In spite of this, we opted to simply measure accuracy on the entire training set, since we achieved better results with that method in preliminary studies.

Finally, we considered two criteria for halting the search process. One could stop adding attributes when none of the alternatives improves classification accuracy, or one could adopt a more conservative strategy of continuing to select attributes as long as they do not degrade accuracy. One argument for the latter approach is that higher dimensional spaces are more likely to allow separation of classes with a single decision boundary, which favors the inclusion of more attributes. Because initial experiments favored this scheme, we incorporated it into the system.

To summarize, the algorithm initializes the subset of attributes to the empty set, and the accuracy of the resulting classifier, which simply predicts the most frequent class, is saved for subsequent comparison. On each iteration, the method considers adding each unused attribute to the subset on a trial basis and measures the performance of the resulting classifier on the training data. The attribute that most improves (or at least maintains) the accuracy is permanently added to the subset, with ties broken randomly. The algorithm terminates when addition of any attribute results in reduced accuracy, at which point it returns probabilistic summaries based on the current attribute set.

## Experiments with Bayesian Classifiers

Previous comparative studies have shown that the naive Bayesian classifier outperforms more sophisticated methods such as decision-tree induction in some domains, but that it performs significantly worse in others (Langley et al., 1992). We hypothesized that the first result reflects decision trees' reliance on axis-parallel splits, which poorly mimic the actual decision boundaries in some domains. In contrast, we posited that the naive Bayesian classifier did poorly in domains containing redundant attributes. Since the selective classifier should not suffer from the latter problem, we predicted that it would improve upon the performance of the naive classifier in the latter domains, perhaps equaling the accuracy of decision-tree methods, while remaining superior in the former domains.

To test this idea, we compared the behavior of the selective Bayesian classifier to that of the naive Bayesian classifier and Quinlan's (1993) C4.5 decision-tree algorithm in six domains from the UCI repository of machine learning databases. We knew that the naive classifier outperforms C4.5 in the soybean disease, breast cancer, and DNA promoter domains, whereas the reverse is true for the mushroom, Congressional voting, and chess endgame domains. Therefore, these domains seemed to provide a good testbed for evaluating the new algorithm.

Each data set contains a set of classified instances described in terms of numeric or nominal attributes. For example, the soybean disease data consists of 47 instances described in terms of climate conditions, crop history, and plant symptoms, each labeled with one of four disease classes. The Congressional voting domain describes the 435 members of the 98th Congress by their votes on 16 key issues and labeled as Democrat or Republican. The breast cancer data includes 699 instances of malignant and benign tissue samples described by nine numeric attributes such as clump thickness, marginal adhesion, and mitoses. Detailed information about these six domains, and many others, is available from the UCI repository via anonymous ftp to ICS.UCI.EDU.

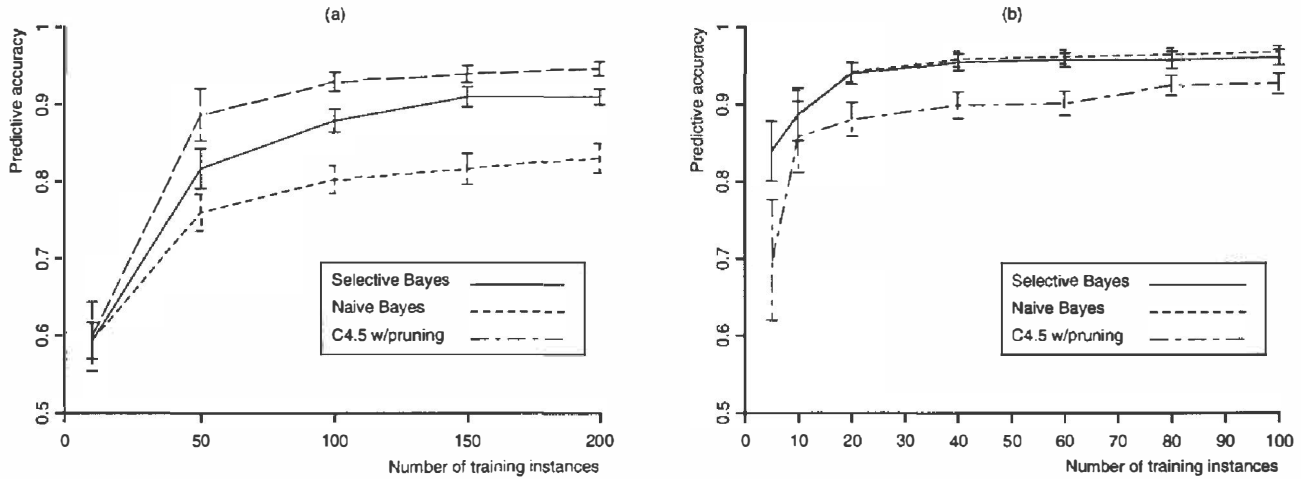

*Figure 2.* Learning curves for the selective Bayesian classifier, the naive Bayesian classifier, and C4.5 with pruning on (a) chess endgames and (b) breast cancer.

For each domain, we randomly generated 20 sets of separate training and test cases. The dependent variable in our experiment was classification accuracy on the test cases after processing a sample of training cases, averaged over the 20 runs. The classification accuracy of an algorithm is the percentage of test cases for which it correctly predicts the class. Since we were interested in the rate of improvement as well as the asymptotic accuracy of the algorithms, we measured accuracy for different numbers of training samples.

Figure 1 (a) and (b) present the resulting learning curves for the Congressional voting and mushroom domains, respectively, with 95% confidence intervals shown for each point. In both cases, asymptotic accuracy for the selective Bayesian classifier is noticeably higher than for the naive method, approaching the level of C4.5 in the voting domain, but remaining slightly lower for the mushroom data. Figure 2 (a) shows an even greater increase in accuracy for the domain of chess endgames, but again the selective classifier does not quite reach the C4.5 level.

Experimental results for the other three domains present a very different picture. Figure 2 (b) shows that the selective algorithm reproduces the superior performance of the naive Bayesian classifier over decision-tree induction in the breast cancer domain. Analogous results appear in Figure 3 (a) and (b) for both the soybean and DNA promoter data. The odd C4.5 behavior on the soybean data occurs with both pruning and non-pruning versions of the program.

These results confirm our predictions about the comparative behavior of the three algorithms. In domains where the naive classifier exhibits low asymptotic accuracy, apparently due to the presence of redundant attributes, the selective Bayesian classifier shows a marked improvement. At the same time, it does as well as the simple classifier in domains where the latter already outperforms decision-tree induction. Thus, the selective Bayesian classifier appears to overcome the weaknesses of the other two algorithms.

## Related Work on Bayesian Induction

Recent years have seen growing interest in probabilistic approaches to induction, and research in this genre has typically followed one of two paths. Briefly, one approach focuses on the introduction of new features and the creation of explicit dependency links, whereas the other emphasizes the clustering of instances into taxonomic hierarchies. Each framework attempts to improve upon the naive Bayesian classifier by extending the basic induction algorithm in significant ways.

Kononenko (1991) describes an example of the first approach that tests for dependencies among attributes and creates new features based on the conjunctions of correlated values. This 'semi-naive Bayesian classifier' uses the training data to compute conditional probabilities for these joint features, using them to classify test cases rather than the original ones. However, experimental comparisons between his algorithm and the naive Bayesian classifier revealed no differences on two medical domains and only slight improvement on two others data sets. Schlimmer's (1987) STAGGER constructed features for analogous reasons and in a similar manner, though it operated within a rather different probabilistic framework.

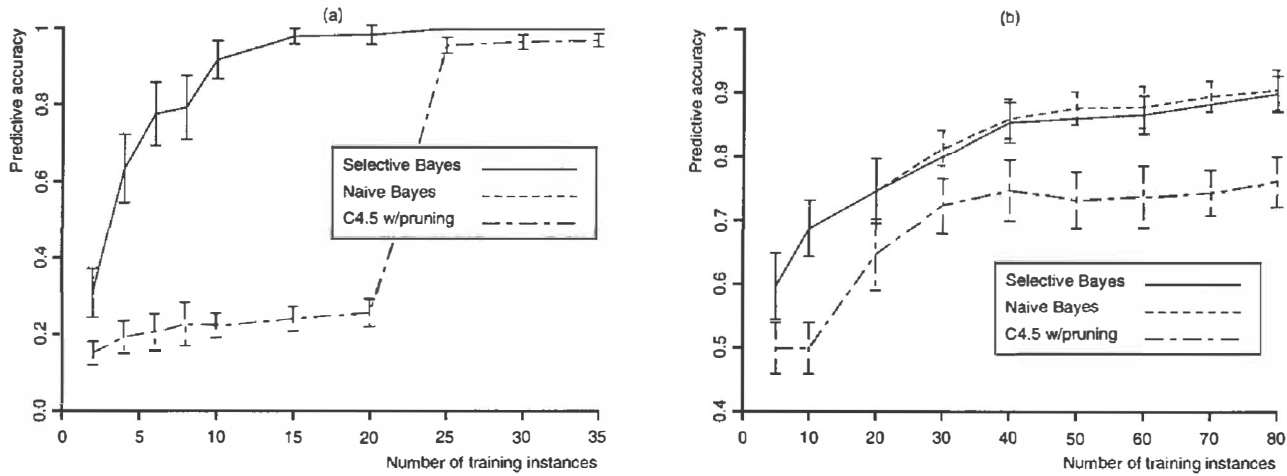

*Figure 3.* Learning curves for the selective Bayesian classifier, the naive Bayesian classifier, and C4.5 with pruning on (a) the small soybean domain and (b) DNA promoters. Selective Bayes incorporates all attributes for the soybean data, giving an identical curve to that for the naive method.

Research on the induction of Bayesian networks (Pearl, 1988) generalizes this basic approach to handling attribute dependence. Cooper and Herskovits' (1992) K2 algorithm carries out a greedy search through the space of Bayesian networks, but it requires the user to specify an ordering on the attributes, and it does not introduce new features. More recently, Connolly (1993) has sidestepped this restriction by using a probabilistic clustering method to generate hidden attributes that render the observable ones conditionally independent. However, only Kononenko has explicitly compared the accuracy of his technique to the naive approach on natural domains, so the usefulness of these methods' increased sophistication remains an open question.

Langley (1993) describes a straightforward example of the hierarchy-building approach. His 'recursive Bayesian classifier' uses the naive algorithm to generate a probabilistic summary for each class. If these summaries correctly classify the training set, the method halts. Otherwise, it calls the naive method recursively for each class to which instances from other classes were assigned, using all cases assigned to that class as training data. The method continues to recurse until it correctly classifies all of the training data or gains no further improvement, then organizes the resulting classifiers as a hierarchy of probabilistic descriptions, which it uses to sort novel test cases. Experiments on artificial domains showed that this algorithm can induce concepts that the naive Bayesian classifier cannot handle, but studies on natural domains showed no significant differences between the methods.

Most work on the induction of probabilistic concept hierarchies builds directly on Fisher's (1987) COBWEB, which deals with unsupervised training data. His incremental algorithm uses an information-theoretic evaluation function to determine when to incorporate a training case into an existing category and when to create an entirely new category. Gennari, Langley, and Fisher (1989), Hadzikadic and Yun (1989), McKusick and Langley (1991), and others have explored very similar approaches. Anderson and Matessa (1992) have adapted the same basic idea within a strict Bayesian framework, though their method creates a flat set of categories rather than a hierarchy. Unfortunately, experiments that compare these clustering schemes to the naive Bayesian classifier are rare, so again one cannot tell whether their sophistication is necessary.

Clearly, the approach we have taken here differs from both of these frameworks for probabilistic induction. Rather than assuming a more sophisticated knowledge structure (and thus requiring more complex methods for using and acquiring that knowledge), the selective Bayesian classifier retains the simplicity of the naive approach but ignores attributes that reduce classification accuracy. We used the assumption of independence to motivate this idea, but it should also prove useful in domains with irrelevant features.

Of course, the basic idea of restricting the attributes used for prediction is not new, nor are greedy approaches for searching the attribute space. Kittler (1986) refers to the scheme we have used as *sequential forward selection* and refers to search in the opposite direction as *sequential backward elimination*. Brodley and Utgoff (1992) have used both methods in their work on multivariate decision trees, whereas John, Kohavi, and Pfleger (in press), Caruana and Freitag (in press), Skalak (in press), and Langley and Sage (in press) have used similar schemes to determine relevant features for decision-tree and nearest neighbor meth-ods. Our contribution lies in extending this idea to Bayesian classifiers, which typically take all attributes into account during prediction.[4]

Superficially, our approach is similar to Michie and Al Attar's (1991) 'sequential Bayesian classifier', which inspects one attribute at a time during classification, selecting the most informative one at each step and halting when the probability of a class exceeds a threshold. However, their method's behavior is better viewed as constructing a decision tree using a probabilistic evaluation function. Our technique has much more in common with the approach reported by Kubat, Flotzinger, and Pfurtscheller (1993), who use decision-tree induction to select predictive attributes for use in a naive Bayesian classifier. They report promising results with this method on an EEG classification task that parallel our findings with the UCI data sets.

## Concluding Remarks

Although our own experimental results have been encouraging, they remain preliminary, and the variety of related approaches suggests many possibilities for additional comparative studies. For example, we should determine the extent to which techniques for inducing Bayesian networks and probabilistic concept hierarchies provide benefits beyond the simple selection scheme we have used here. We should also carry out more systematic studies to explore the effect of the design decisions we made when implementing the selective Bayesian classifier.

In addition, we should consider the usefulness of other selection techniques, such as Kubat et al.'s method, and compare our technique to frameworks with similar representational power that do not rely on the independence assumption, such as the LMS algorithm and related techniques (Widrow & Winter, 1988). The simplicity of the selective Bayesian classifier should also lend itself to average-case analyses (Langley et al., 1992), which would let us compare our experimental results to theoretical ones, at least in synthetic domains.

In summary, we found that a simple modification to the naive Bayesian classifier – forward selection of attributes using estimated accuracy – increases asymptotic accuracy on separate test sets in some domains and does not harm accuracy in others. The selection algorithm appears to be beneficial in domains that involve significant correlations among the predictive attributes, which can bias the decisions of the naive Bayesian classifier if they are not removed. The result is a technique that improves on an already robust algorithm, and that extends the repertoire of methods for probabilistic induction.

## Acknowledgements

We owe thanks to George John, Ronny Kohavi, Igor Kononenko, Karl Pfleger, Jeff Schlimmer, and Miroslav Kubat for discussions on various ideas reported in this paper and to Ray Mooney, who provided modified C4.5 code for producing learning curves. The Stanford University Robotics Laboratory provided space and facilities that greatly aided our research. This work was supported in part by a grant from the Office of Naval Research. The review of the naive Bayesian classifier in Section 2 repeats some material from Langley (1993).

---

4. Warner, Toronto, Veasy, and Stephenson (1961) presented one of the earliest arguments in favor of removing correlated features from the naive Bayesian classifier, but they carried out this process manually.